\documentclass[sigconf]{acmart}
\setcopyright{none}
\usepackage{algorithm}
\usepackage{algorithmic}
\usepackage{tabularx} 
\usepackage{multicol} 
\usepackage{multirow}
\usepackage{hyperref}
\usepackage{graphicx}

\AtBeginDocument{%
  }

\copyrightyear{2025} 
\acmYear{2025} 
\setcopyright{acmlicensed}
\setcopyright{rightsretained}
\acmConference[WWW Companion '25] {Companion Proceedings of the ACM Web Conference 2025}{April 28-May 2, 2025}{Sydney, NSW, Australia.}
\acmBooktitle{Companion Proceedings of the ACM Web Conference 2025 (WWW Companion '25), April 28-May 2, 2025, Sydney, NSW, Australia}
\acmISBN{979-8-4007-1331-6/25/04} 
\acmDOI{10.1145/3701716.3735084}
\begin{document}


\title{HALO: Half Life-Based Outdated Fact Filtering in Temporal Knowledge Graphs}






 \author{{Feng Ding}}
\affiliation{%
 \institution{Dalian University of Technology}
 \city{Dalian}
 \country{China}
 }
  \email{dingfeng@dlut.edu.cn}

\author{Tingting Wang}
\affiliation{%
 \institution{Dalian University of Technology}
 \city{Dalian}
 \country{China}
 }
 \email{ttwang93@outlook.com}

\author{Yupeng Gao}
\affiliation{%
 \institution{Dalian University of Technology}
 \city{Dalian}
 \country{China}
 }
  \email{commer@mail.dlut.edu.cn}

\author{Shuo Yu}
\authornote{Corresponding author}
\affiliation{%
 \institution{Dalian University of Technology,}
 \institution{Key Laboratory of Social Computing and Cognitive Intelligence, Ministry of Education}
  \city{Dalian}
  \country{China}
 }
 \email{shuo.yu@ieee.org}

\author{Jing Ren}
\authornotemark[1]
\affiliation{%
  \institution{RMIT University}
  \city{Melbourne}
  \country{Australia}
}
\email{jing.ren@ieee.org}

\author{Feng Xia}

\affiliation{%
 \institution{RMIT University}
 \city{Melbourne}
 \country{Australia}
 }
 \email{f.xia@ieee.org}

\renewcommand{\shortauthors}{Ding et al.}

\begin{abstract}
  Outdated facts in temporal knowledge graphs (TKGs) result from exceeding the expiration date of facts, which negatively impact reasoning performance on TKGs. However, existing reasoning methods primarily focus on positive importance of historical facts, neglecting adverse effects of outdated facts. Besides, training on these outdated facts yields extra computational cost. To address these challenges, we propose an outdated fact filtering framework named HALO, which quantifies the temporal validity of historical facts by exploring the half-life theory to filter outdated facts in TKGs. HALO consists of three modules: the temporal fact attention module, the dynamic relation-aware encoder module, and the outdated fact filtering module. Firstly, the temporal fact attention module captures the evolution of historical facts over time to identify relevant facts. Secondly, the dynamic relation-aware encoder module is designed for efficiently predicting the half life of each fact. Finally, we construct a time decay function based on the half-life theory to quantify the temporal validity of facts and filter outdated facts. Experimental results show that HALO outperforms the state-of-the-art TKG reasoning methods on three public datasets, demonstrating its effectiveness in detecting and filtering outdated facts\footnote{Codes are available at \url{https://github.com/yushuowiki/K-Half/tree/main}}.
  \end{abstract}





\maketitle

\section{Introduction} \label{Intro}
Temporal knowledge graphs (TKGs) serve as a powerful tool for modeling dynamic real-world data, which have achieved significant success in various domains, such as question answering \cite{DBLP:conf/aaai/XueLWZ24},
recommendation system \cite{DBLP:conf/www/ZhangSYLX25},
and knowledge retrieval \cite{DBLP:conf/emnlp/QianZZZSZS24}.
Temporal knowledge graph reasoning (TKGR) aims at predicting future events based on the historical facts within TKGs. During this process, outdated facts in TKGs can have several negative impacts across different domains, such as poor recommendations and user experience,  biased AI models, and ethical issues.
Despite prior TKGR studies performing well by capturing the evolution of knowledge through historical facts, leveraging repeating patterns~\cite{han2020explainable}, cyclic trends~\cite{DBLP:conf/aaai/ZhuCFCZ21}, and temporal dependencies~\cite{DBLP:conf/sigir/LiJLGGSWC21}, they primarily focus on the positive importance of historical facts while neglecting the negative impact of outdated information~\cite{tu2023deep}.
In addition, training on outdated facts requires extra computation cost, bringing significant challenges in the efficiency of model training.
Therefore, effectively filtering outdated facts within TKGs is significant for improving the performance and efficiency of TKGR.

Given the dynamic nature of TKGs, the facts are continuously updated and lead to a decrease in their temporal validity during TKGR.
As illustrated in Figure \ref{fig:intro}, the temporal validity of facts within a TKG is changing over time.
From observations, the temporal validity of certain historical facts falls dramatically after exceeding a certain time interval.
This time interval is similar to the half life when a substance decays to half of its initial quantity.
Inspired by this observation, we employ the half-life theory
to filter outdated facts for enhancing the performance of TKGR tasks.
If the duration time of a fact exceeds its half-life, the fact becomes an outdated fact and should be filtered out the TKGs.
Additionally, different types of facts might have various half-lives.
For example, active facts that update more frequently usually have a short half-life.
It is also a challenge to categorize the facts within TKGs into active facts and inactive facts.
Therefore, leveraging various temporal patterns and temporal dependencies from the two types of facts shows as a promising way for effectively filtering outdated facts.

To tackle the above challenges, we propose a \textbf{HA}lf \textbf{L}ife-based \textbf{O}utdated Facts Filtering framework (\textbf{HALO}).
HALO effectively quantifies the temporal validity of historical facts based on the half-life theory to filter outdated facts in TKGs.
Specifically, HALO consists of three modules: the temporal fact attention module, the dynamic relation-aware encoder module, and the outdated fact filtering module.
In the first module, HALO captures the evolution of historical facts based on a temporal attention-based mechanism, which integrates the temporal information into representations to identify relevant facts within TKGs.
In the dynamic relation-aware module, HALO designs a classifier to identify active or inactive facts within TKGs, and predicts various half-lives of different facts. 
Finally, we construct a time decay function based on the half-life to quantify temporal validity of facts, and outdated facts are filtered by evaluating the probability of temporal validity for each TKG fact.
To verify the effectiveness of HALO, we conduct experiments on the state-of-the-art TKGR methods.
In summary, this work presents the following contributions:

\begin{figure}
    \centering
    \includegraphics[width=\linewidth]{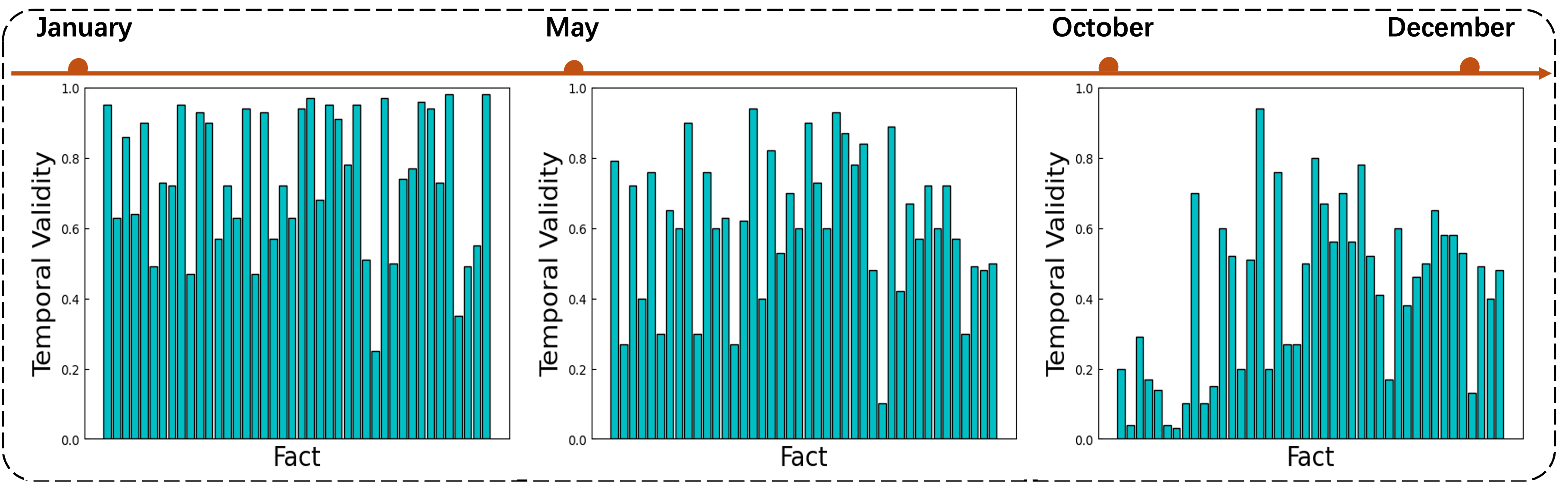}
    \caption{The changing temporal validity of facts within ICEWS14 dataset during the whole year of 2014.}
    \label{fig:intro}
\end{figure}

\begin{itemize}

    
    \item  
    We propose HALO, an efficient framework for filtering outdated facts to improve the performance of TKGR, which is capable of capturing evolution of historical facts and quantifying temporal validity of facts within TKGs.
    
    \item 
    We design a temporal fact attention module to extract temporal and structural dependencies among historical facts and a dynamic relation-aware module based on a facts-to-nodes graph to predict the half life of each fact.
    An outdated fact filtering module based on a time decay function is utilized to effectively quantify the temporal validity of facts and filter outdated facts.
    
    \item 
    We conduct extensive experiments on three real-world TKGs, including ICEWS14, ICEWS18 and ICEWS05-15.
    The experimental results demonstrate that our framework outperforms the state-of-the-art TKGs reasoning baselines.
\end{itemize}

\section{HALO: An Outdated Fact Filtering Framework}
\begin{figure*}
    \centering
    \includegraphics[width=15cm, height=5cm, keepaspectratio]{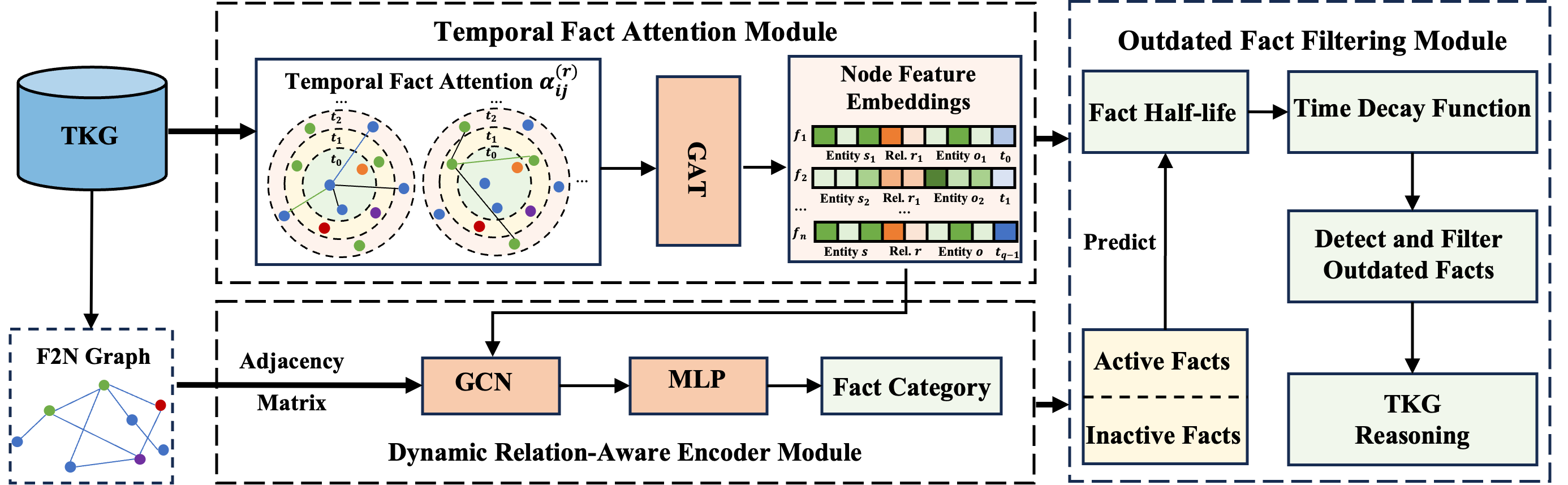}
    \caption{Framework of HALO.}
    \label{fig:framework}
\end{figure*}

In TKGs, outdated facts~\cite{tu2023deep} refer to facts whose duration date exceeds their expiration date.
The outdated fact filtering framework named HALO is shown in Fig.\ref{fig:framework}.
Specifically, we design a temporal fact attention module and a dynamic relation-aware encoder module to learn temporal and structural representations of facts and categorize them into active facts and inactive facts through their representations.
After obtaining the categories of facts in the former modules, HALO predicts the half-life of facts based on their activity such as active facts or inactive facts.
To further quantify the temporal effectiveness of historical fact, HALO designs a time decay function based on the predicted half-life of the fact.
Finally, HALO can efficiently filter outdated facts to improve the performance of TKGR tasks by leveraging the time decay function. 

\subsection{Temporal Fact Attention Module}
Before starting to learn the features, the features of entities and relations are randomly initialized as $\mathbf{E}=\{\mathbf{e}_1,\mathbf{e}_2,...,\mathbf{e}_{n1}\}\in \mathbb{R}^{n_1 \times d_1}$ and \(\textbf{R} = \{\textbf{r}_1,\textbf{r}_2, ..., \textbf{r}_{n_2} \} \in \mathbb{R}^{n_2 \times d_2}\), 
where $\mathbf{e}_i$ and \(\textbf{r}_i\) are the feature vector,
$n_1=|\mathcal{E}|$ and \(n_2 = |\mathcal{R}|\) denote the total number of entities and relations, and $d_1$ and \(d_2\) represent the dimension of feature vector.
To integrate the temporal information into representations, we encode the specific update time interval $d$ and concatenate it into the representation of the fact.
Taken the fact $f_{ij,t_i}^{(r_x)}$ that contains the head entity $e_i$ and the tail entity $e_j$ with relation $r_x$ at timestamp $t_i$ as an example, its embedding $\boldsymbol{f}_{ij,t_i}^{(r_x)}$ can be formulated according to the following equation:
\begin{equation}
    \boldsymbol{f}_{ij,t_i}^{(r_x)}=
    \mathbf{W}_1 [\mathbf{e}_i||\mathbf{r}_x||\mathbf{e}_j||\phi(d)],\phi(d) = cos (d\omega_t+b_t),
\end{equation}
where $d=t_c-t_i$ is the update time interval between the current time $t_c$ and the update time $t_i$, $\phi(d)$ is modeled by rescaling a learnable time unit $\omega_t$ with a time bias $b_t$, and $cos(\cdot)$ is a periodic activation function. $\mathbf{W}_1$ is a learnable parametric matrix.
The attention score $\alpha_{ij}^{(r_x)}$ to the fact with the head entity ${e_i}$, tail entity $e_j$ and relation $r_x$ at timestamp $t_i$ can be calculated as follows:
\begin{align}
    \alpha_{ij}^{(r_x)}  &=  softmax(\sigma(\boldsymbol{W}_2\boldsymbol{f_{ij,t_i}^{(r_x)}})), 
\end{align}
where $\sigma$ denotes parametric LeakyReLU function.
$\mathbf{W}_2$ is a learnable parametric matrix.
To stabilize the learning process, we use \textit{multi-head attention} mechanism, and the final  entity embedding $\mathbf{e}_i^{\prime} \in \mathbb{R}^{d_1^{\prime}}$ of the entity $e_i$ is expressed as follows:
\begin{equation}
     e_{i}^{\prime} = \sigma \left( \frac{1}{M} \sum_{m=1}^{M}
     \sum_{j\in \mathcal{N}_i} \sum_{x \in \mathcal{R}_{ij},t\in \tau} 
    a_{ij}^{(r_x),m}f_{ij,t_i}^{(r_x),m} \right),
\end{equation}
where 
$M$ is the total number of heads, $\tau$ is the total number of timestamps, $i$ is the index of head entity $e_i$, $j$ is the index of tail entity $e_j$, and $x$ is the index of relation $r_x$. 
And we obtain the embeddings $\mathbf{F}=\{\mathbf{F}_{r_1},\mathbf{F}_{r_2},...,\mathbf{F}_{r_{n_2}}\} \in \mathbb{R}^{n_3 \times (2d_1+d_2+1)}$ of the temporal facts, where $n_3$ is the number of facts. 

We define the loss function as:
\begin{equation}
    \mathcal{L}_{gat} = 
    \sum_{f_{ij}\in F}
    \sum_{f_{ij}^{\prime}\in F^{\prime}}
    max(
       d_{f_{ij}^{\prime}} -
       d_{f_{ij}} +
       \theta,0
    ),
\end{equation}
where $f_{ij} \in F$ denotes the set of valid facts, $f_{ij}^{\prime} \in F^{\prime}$ represents the set of invalid facts by replacing the head $e_i$ or tail $e_j$ entity of a quadruple based on $F$, and $\theta>0$ is a hyperparameter to fine-tune the weight of this layer.
$d_{f_{ij}}=\rVert\mathbf{e}_i + \mathbf{r}_x -\mathbf{e}_j \lVert_1$ represents the embedding distance
using the dissimilarity measure of the L1 norm.
Similarly, $d_{f_{ij}^{\prime}}$ denotes the distance between the invalid facts.

\subsection{Dynamic Relation-aware Encoder Module}
In the former module, we have obtained the fact embeddings $\mathbf{F}$
based on the sharing head entities of facts.
However, the activity of each fact is related not only to its head entity but also to its relation.
Thus, we design a dynamic relation-aware encoder to capture more expressive representations of temporal facts. 
Firstly, we construct a facts-to-nodes graph $G_{F2N}=\{G_{r_1},G_{r_2},...,G_{r_{n_2}}\}$ to take into account all facts with the same relation such as $r_i\in R$.
In the graph $G_{F2N}$, the nodes are transformed from facts within the TKGs.
If two facts hold the same relation, the edge weight between them will increase to one.
Then, we perform a linear transformation on $G_{F2N}$ through a shared learnable weight matrix $\mathbf{W}_3$.
Taken the facts with the same relation $r_x$ as an example, the final fact embeddings $\mathbf{\hat{F}}_{r_x}$ with relation $r_x$ is further expressed as follow:
\begin{equation}
    \mathbf{\hat{F}}_{r_x} = \mathbf{F}_{r_x} \cdot \mathbf{W}_3,
\end{equation}
where $\mathbf{\hat{F}_{r_x}}\in \mathbb{R}^{n_{r_x} \times d_3^{\prime}}$, 
$n_{r_x}$ denotes the total number of facts with relation $r_x$,
and $d_3^{\prime}$ represents the final embedding dimension of facts.
Furthermore, we use a multi-layer graph convolution network as the encoder, and the message propagation mechanism can be formulated as follows:
\begin{equation}
    \textbf{H}^{(l)}_{r_x} = \sigma \left( \hat{\textbf{D}}^{-\frac{1}{2}} \hat{\textbf{A}}_{r_x} \hat{\textbf{D}}^{-\frac{1}{2}} \textbf{H}^{(l-1)}_{r_x} \boldsymbol{\Theta}^{(l-1)} \right),
\end{equation}
where \(\hat{\textbf{A}}_{r_x} = \textbf{A}_{r_x} + \textbf{I}_N\) is an adjacency matrix $\mathbf{A}_{r_x}\in \mathbb{R}^{n_{r_x} \times n_{r_x}}$ of graph $G_{r_x}$ with inserted self-loop and \(\hat{\textbf{D}}\) denotes the degree of matrix \(\hat{\textbf{A}}_{r_x}\).

Then, we add a fully connected layer after the relation-aware dynamic encoder to categorize facts into \textit{active facts} $\mathcal{F}_{act}$ and \textit{inactive facts} $\mathcal{F}_{ina}$ for predicting their half-lives.
Since we formulate the fact category prediction as a binary classification task, we use the BEC loss to train the fact type classifier model.
The objective function is expressed as:
\begin{equation}
    \mathcal{L}_{bec} = -[ylog\hat{y}+(1-y)log(1-\hat{y})],
\end{equation}
where $\hat{y}$ denotes the predicted label, and $y$ represents the true label. 
If a fact is active, we set $y=0$. Otherwise, $y=1$.

\subsection{Outdated Fact Filtering Module}
Then, HALO will predict their half-life based on the update information of the active facts or inactive facts. 
\subsubsection{Fact Half-life Prediction}
In a TKG, if the tail entity of relation $r_x$ connected to the head entity $e_i$ is changed from $e_j$ to $e_k$ at timestamp $t_j$, it is considered as one update operator of fact.
And one update time interval is expressed as $\Delta t = t_j - t_i$, where $t_j > t_i$.
In reality, since a fact $f_i^{(r_x)}$ with head entity $e_i$ and relation $r_x$ can be updated many times, we use the average update time interval as the final update time interval $\Delta t$.
In this paper, we assume that active facts $\mathcal{F}_{act}=\{f_0,f_1,...,f_{K-1}\}$ have the similar update time interval and share the same half-life, where $K$ is the total number of active facts.
Thus, the half-life of active facts is expressed as follows:
\begin{equation} \label{equation_11}
    t_{HF_{act}}=
    \frac{1}{2}
    \cdot
    \frac{\sum_{i=0}^{K-1}\Delta t_{f_{i}\in \mathcal{F}_{act}}}{K}.
\end{equation} 
If the previous fact was updated by the new fact, its validity becomes 0, and the validity of the new fact equals to 1.
The half-life of a fact represents the time at which the validity of it decreased to half of its initial value.
This is the reason why we add $\frac{1}{2}$ to Eq. \eqref{equation_11}.

Similar to predict half-life of active facts, the half-life of inactive facts is formulated as active facts.
Thus, the half-life of each fact in a TKG can be expressed as follow:
\begin{equation}
    t_{HF} = 
    \begin{cases}
        t_{HF_{act}}, & \text{if $f_i\in \mathcal{F}_{act}$ } \\
        t_{HF_{ina}}, & \text{if $f_i\in \mathcal{F}_{ina}$ }
    \end{cases}.
\end{equation}

\subsubsection{Outdated Fact Filtering and Verification}
Given a fact $f=(s,r,o,t_i)$ with a half-life $t_{HF}$ as an example, the temporal effectiveness of the fact is expressed as follows:
\begin{equation} \label{e.v_f}
    V(t_i,t_{HF})=V_0 \cdot
    e^{- \lambda \cdot (t_c-t_i)}, 
    \lambda = \frac{ln(2)}{t_{HF}},
\end{equation} 
where $t_c \geq t_i$, $V_0$ is the initial influence and it is set to 1 in this work.
In addition, $\lambda$ denotes the decay rate of a fact and is derived from the half-life of the fact.
When the influence of fact $V_f(t_i,t_{Hf})$ falls below a predefined threshold $\theta$, the fact turns out to be an outdated fact and should be filtered from the set $\mathcal{F}$ in a TKG.
Furthermore, the expiration time can be inferred from Eq.\eqref{e.v_f}.

The final loss function is computed as:
\begin{equation}
    \mathcal{L}=\alpha\mathcal{L}_{gat}+(1-\alpha)\mathcal{L}_{bec},
\end{equation}
where $\alpha \in [0,1]$ is a hyper-parameter that controls the influence of the module.

\section{Experiments}
We conduct experiments on three widely used datasets: ICEWS14, ICEWS18, ICEWS05-15.
We use state-of-the-art reasoning tasks, such as CyGNet \cite{DBLP:conf/aaai/ZhuCFCZ21}, xERTE \cite{han2020explainable}, RE-GCN \cite{DBLP:conf/sigir/LiJLGGSWC21}, 
CENET \cite{DBLP:conf/aaai/XuO0F23} and LogCL \cite{DBLP:conf/icde/ChenWWZCLL24}, to verify the negative impacts of outdated facts in TKGs.
Besides, we adopt two evaluation metrics: mean reciprocal rank (MRR) and Hits@1, which are widely used to evaluate the effectiveness of TKGR methods. 

\begin{table}[htp]
\centering
\begin{tabular}{ccccccccccccc}
\toprule
 & \multicolumn{2}{c}{\textbf{ICEWS14}} & \multicolumn{2}{c}{\textbf{ICEWS18}} & \multicolumn{2}{c}{\textbf{ICEWS05-15}}  \\ \cline{2-7} 
\multirow{-2}{*}{\textbf{Model}} & \multicolumn{1}{c}{MRR} & \multicolumn{1}{c}{H@1} 
& \multicolumn{1}{c}{MRR} & \multicolumn{1}{c}{H@1} 
& \multicolumn{1}{c}{MRR} & \multicolumn{1}{c}{H@1} 
\\ \hline

{CyGNet} & 45.45 & 37.46  & 44.45 & 37.59  & 54.27 & 46.64  \\ 

{+HALO} & 47.88 $\uparrow$ &  41.24$\uparrow$ &  44.72$\uparrow$ & 37.99$\uparrow$ & 56.02$\uparrow$ & 48.48$\uparrow$  \\ \hline

{xERTE} & 40.74 & 32.68  & 29.03 & 20.69 
&46.47 &37.73   \\

{+HALO} & 41.13$\uparrow$ &  33.09$\uparrow$ 
& 30.1$\uparrow$ &  21.77$\uparrow$  &  46.48$\uparrow$ & 38.00$\uparrow$  \\ \hline

{RE-GCN} &37.67	&27.68	&27.74	&17.94	&37.56	&26.73 \\ 

{+HALO} & 41.38$\uparrow$	& 30.52$\uparrow$	& 30.43$\uparrow$	& 19.91$\uparrow$& 38.12$\uparrow$	& 27.53$\uparrow$ \\ \hline

{CENET} &47.05	&\underline{43.40}	&\underline{45.89}	&\underline{42.78}&\underline{63.62}	&\underline{60.71} \\

{+HALO} &47.79$\uparrow$	&\textbf{44.17}$\uparrow$ &\textbf{46.06}$\uparrow$	&\textbf{42.86}$\uparrow$	&\textbf{65.73}$\uparrow$	&\textbf{62.83}$\uparrow$\\ \hline

{LogCL} & \underline{48.00}	&{37.00}	&35.67	&24.53	&57.04	&46.07\\

{+HALO} & \textbf{48.20}$\uparrow$	&37.20$\uparrow$ &36.11$\uparrow$	&24.7$\uparrow$	&58.12$\uparrow$	&46.98$\uparrow$ \\
 \bottomrule
\end{tabular}
\caption{Performance (in percentage) for reasoning over TKGs on ICEWS14, ICEWS18, and ICEWS05-15 datasets.}
\label{tab:my-table_1}
\end{table}

\begin{figure}
    \centering
    \includegraphics[width=\linewidth]{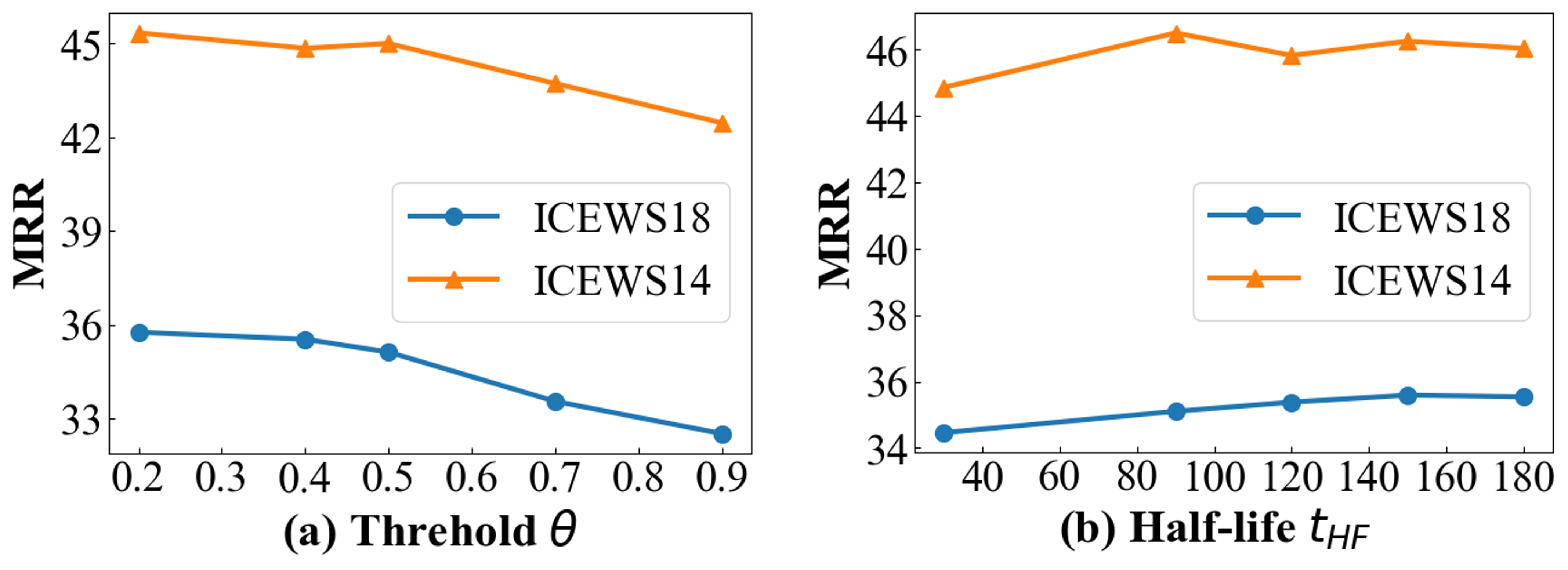}
    \caption{Sensitivity analysis on ICEWS18 and ICEWS14.}
    \label{fig:enter-label}
\end{figure}

\subsection{Results of TKG Reasoning}
HALO (ours) filters outdated facts on ICEWS14, ICEWS18, and ICEWS05-15 datasets.
To verify our assumption that outdated facts actually have a negative impact on the performance of TKGs reasoning, we conduct reasoning experiments on these datasets. 
The general experimental results are reported in Table \ref{tab:my-table_1}. 
Reasoning methods adding HALO (i.e. +HALO) represents conducting experiments on the datasets that have filtered outdated facts through HALO. 
The best results are marked in bold, and the second-best results are reported using underline. 
From the table, we observe that no matter which TKG reasoning approaches we compare, filtering outdated facts with HALO always improves performance when predicting future events.

\subsection{Sensitivity Analysis}
We conduct experiments on ICEWS14 and ICEWS18 datasets to further analyze the impact of parameters in HALO, including the parameters $\theta$ and $t_{HF}$.
To explore the influence of parameter $\theta$, we conduct experiments with $\theta$ from 0 to 1 while other optimal hyperparameters fixed.
The results of MRR are shown in Figure \ref{fig:enter-label} (a).
A higher value of $\theta$ indicates that a lower proportion of facts are out of date.
It can be seen that the performance of LogCL+HALO shows a trend of falling as $\theta$ increases on ICEWS14 and ICEWS18, indicating that a lower influence of a fact is more likely to be outdated.
Furthermore, it is consistent with our assumption.
As mentioned previously, active facts are more likely to become outdated.
To verify the half-life of the fact is related to its historical update information, we perform a batch of experiments with $t_{HF}$ being randomly set.
Different datasets are affected by various half-life, which indicates that different historical update information results in different half-life of fact. 
As a number of facts in the ICEWS14 datasets are active facts, the performance of reasoning tasks achieves the best results with a half-life around 80 days.
In contrast, the facts in the ICEWS18 datasets are usually inactive, so the performance of reasoning tasks achieves the best results with a half-life around 160 days.

\section{Conclusion}

In this paper, we introduced HALO, a novel half-life-based outdated fact filtering framework designed to enhance TKGR by addressing the overlooked issue of outdated facts. By quantifying the temporal validity of facts, HALO ensures that TKGs evolve dynamically and maintain relevance over time. Experimental results on benchmark datasets have demonstrated that HALO significantly improves reasoning accuracy by effectively filtering outdated information, leading to more precise temporal predictions. Additionally, our analysis highlights the importance of considering fact expiration in TKGR models to avoid misleading inferences caused by obsolete knowledge. 
In future work, we will focus on expanding the framework to handle multi-modal KGs, incorporating data from different sources like text, images, or sensor data, for improving the robustness and versatility of temporal reasoning systems. 

\section*{Acknowledgments}
This work is partially supported by National Natural Science Foundation of China under Grant No. 62102060 and the Fundamental Research Funds for the Central Universities under Grant No. DUT24LAB121.

\bibliographystyle{ACM-Reference-Format}
\bibliography{sample-base}










\end{document}